\documentclass[lettersize,journal]{IEEEtran}
\usepackage{amsmath,amsfonts}
\usepackage{autobreak}
\usepackage{algorithmic}
\usepackage{algorithm}
\usepackage{array}
\usepackage[caption=false,font=normalsize,labelfont=sf,textfont=sf]{subfig}
\usepackage{textcomp}
\usepackage{stfloats}
\usepackage{url}
\usepackage{verbatim}
\usepackage{graphicx}
\usepackage{cite}
\usepackage{hyperref}
\usepackage{graphics}
\usepackage{epsfig}
\usepackage{gensymb}
\usepackage{algorithm} 
\usepackage{algorithmic} 
\usepackage{multirow} 
\usepackage{utfsym}
\usepackage{tabularx}
\usepackage{booktabs}
\usepackage{color}
\usepackage{makecell}
\usepackage{soul, color, xcolor}
\begin{document}

\title{CAD-Mesher: A Convenient, Accurate, Dense Mesh-based Mapping Module in SLAM for Dynamic Environments}

\author{Yanpeng Jia, Fengkui Cao$^{*}$, Ting Wang$^{*}$, Yandong Tang, Shiliang Shao and Lianqing Liu
\thanks{$^{*}$This work was supported by National Natural Science Foundation of China (Grant No. 62203091 and U20A20201), the China postdoctoral Science Foundation (Grant No. GZB20230804), Autonomous Project of State Key Laboratory of Robotics (Grant No. 2024-Z09). \textit{(corresponding author: Ting Wang, Fengkui Cao)}}
\thanks{Y. Jia, F. Cao, T. Wang, Y. Tang, S. Shao and L. Liu are with the State Key Laboratory of Robotics at Shenyang Institute of Automation, Chinese Academy of Sciences, Shenyang, China. Y. Jia is also with the University of Chinese Academy of Sciences, Beijing, China.(email: \{jiayanpeng, caofengkui, wangting, tangyandong, shaoshiliang, liulianqing\}@sia.cn)}}

\markboth{Journal of \LaTeX\ Class Files,~Vol.~14, No.~8, August~2021}%
{Shell \MakeLowercase{\textit{et al.}}: A Sample Article Using IEEEtran.cls for IEEE Journals}

\IEEEpubid{0000--0000/00\$00.00~\copyright~2021 IEEE}

\maketitle

\begin{abstract}
Most LiDAR odometry and SLAM systems construct maps in point clouds, which are discrete and sparse when zoomed in, making them not directly suitable for navigation. Mesh maps represent a dense and continuous map format with low memory consumption, which can approximate complex structures with simple elements, attracting significant attention of researchers in recent years. However, most implementations operate under a static environment assumption. In effect, moving objects cause ghosting, potentially degrading the quality of meshing. To address these issues, we propose a plug-and-play meshing module adapting to dynamic environments, which can easily integrate with various LiDAR odometry to generally improve the pose estimation accuracy of odometry. In our meshing module, a novel two-stage coarse-to-fine dynamic removal method is designed to effectively filter dynamic objects, generating consistent, accurate, and dense mesh maps. To our best know, this is the first mesh construction method with explicit dynamic removal. Additionally, conducive to Gaussian process in mesh construction, sliding window-based keyframe aggregation and adaptive downsampling strategies are used to ensure the uniformity of point cloud. We evaluate the localization and mapping accuracy on five publicly available datasets. Both qualitative and quantitative results demonstrate the superiority of our method compared with the state-of-the-art algorithms. The code and introduction video are publicly available at \url{https://yaepiii.github.io/CAD-Mesher/}.
\end{abstract}


\section{Introduction} \label{sec1}
\IEEEPARstart{I}{n} recent years, 3D LiDAR odometry (LO) and SLAM have been extensively studied, effectively improving localization accuracy and mapping quality. Most LO systems typically generate point cloud maps \cite{survey,d-liom}. However, the memory cost also grows significantly with the increased scale or density. Although point cloud maps seem dense to human eyes, it would still be discrete and sparse when zoomed in \cite{slamesh}. Consequently, point cloud maps often require post-processing, such as voxelization, converted into a continuous map for navigation.

\begin{figure}[tbp]
	\centering
	\includegraphics[width=8.5cm]{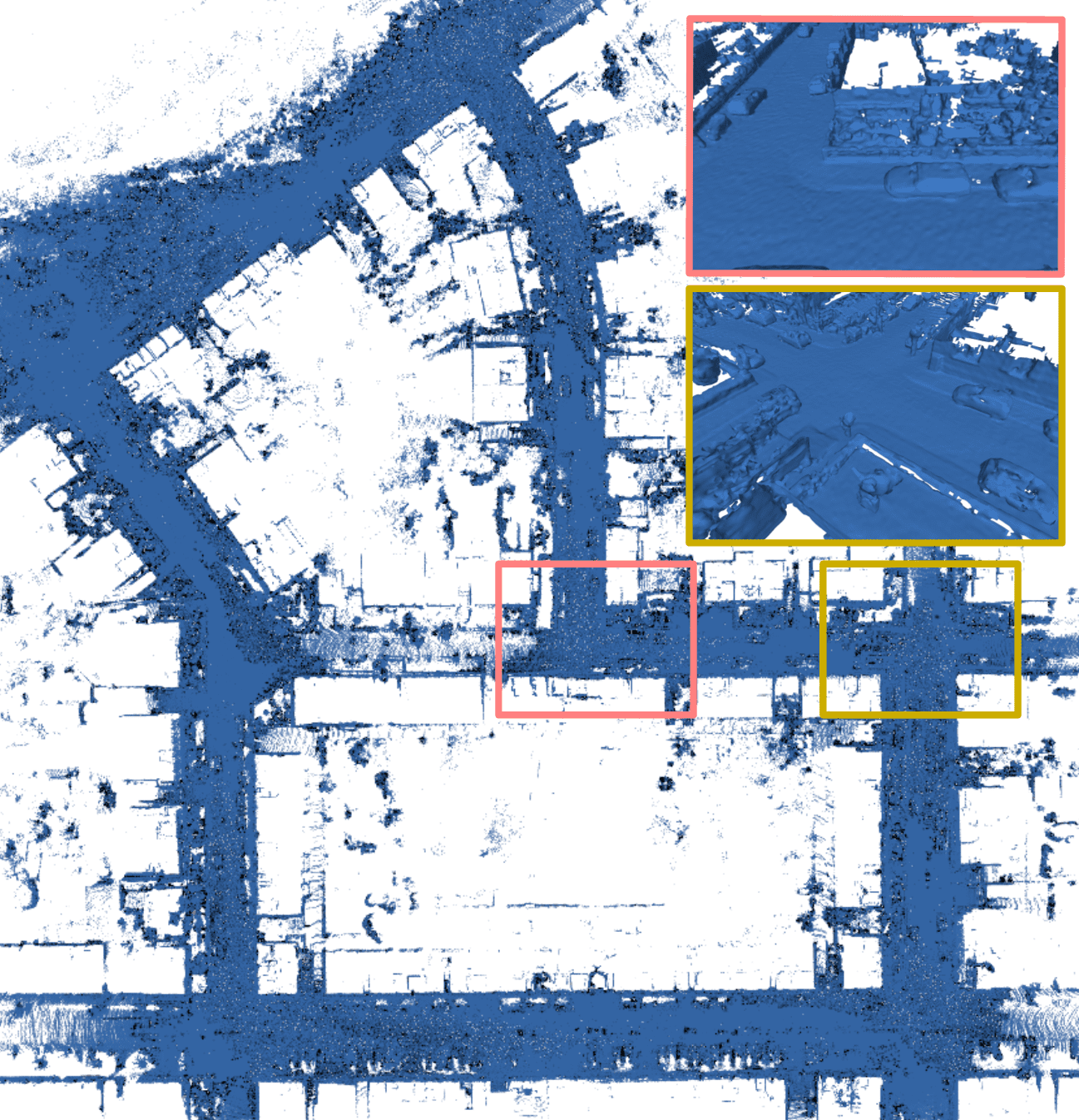}
	\caption{The global mesh map constructed by CAD-Mesher on the KITTI07 sequence. The zoomed-in regions (highlighted by red and yellow boxes) demonstrate the good performance of our method on presenting sufficient details.}
	\label{figure1}
\end{figure}

Therefore, many researchers have been pursuing more compact and denser maps, which facilitate seamless integration with downstream tasks \cite{immesh}. Some methods utilize the Normal Distribution Transform (NDT) \cite{ndt-loam,litamin,litamin2} or surfel \cite{suma,suma++,mars} to parameterize the environment maps, leading to gaps between elements, which limits the ability to capture finely environmental details. Other methods divide the 3D space into voxels using occupancy grid \cite{octomap,ufomap} or truncated signed distance function (TSDF) \cite{vdbfusion}. Although these representations are suitable for navigation, they are not helpful for registration and robot localization, and do not contribute to constructing a map in sufficient detail.

\IEEEpubidadjcol
Since complex 3D structures can be approximated with simple formats, triangular mesh has become a popular representation \cite{vsb}. The triangular mesh representation not only offers high memory efficiency and scalability, but also constructs continuous and smooth surfaces, providing valuable information for registration \cite{rangeimage-based,micp-l,slamesh}. Puma. \cite{puma} use the Poisson surface reconstruction method to achieve smooth mesh construction for outdoor scenes. However, its performance does not meet real-time requirements. Some of the latest methods \cite{shine-mapping,mesh-loam} have succeeded in constructing large-scale 3D mesh maps, but they require extra GPU resources. SLAMesh \cite{slamesh} incrementally builds the mesh maps using Gaussian process (GP), allowing odometry and meshing to mutually benefit each other. However, it requires to balance between efficiency and the precision of mesh map. Additionally, the aforementioned methods assume a primarily static environment and do not filter the impact of dynamic objects, which may lead to registration inconsistencies and meshing degradation.

In this work, we propose a novel meshing module adapting to dynamic environments, CAD-Mesher. Inspired by the plug-and-play characteristics of deep learning network layer modules \cite{strongsort}, CAD-Mesher is designed for easily integrating with various LiDAR odometry (only need to remap the point cloud topic and the odometry topic, without requiring any other modifications), both further improving the accuracy of odometry and incrementally constructing an accurate, dense, and continuous mesh map (as shown in Fig.~\ref{figure1}). Specifically, we propose a two-stage coarse-to-fine dynamic removal approach. Before registration, dynamic points are coarsely removed using a visibility-based method to improve the accuracy of point-to-mesh registration. After registration, a voxel-based probabilistic method is used to finely remove dynamic points, ensuring the quality of meshing. Additionally, we use a keyframe sliding window to accumulate scans of spatial neighbors to provide more accurate training points for GP and achieve denser meshing. To improve efficiency, adaptive downsampling and keyframe selection strategies are introduced to evenly downsample the accumulated point clouds. The main contributions of our work are summarized as follows:

\begin{itemize}
	\item An accurate, dense, and continuous meshing module adapting to dynamic environments, which is plug-and-play for conveniently integrating with various LiDAR odometry to further improve the pose estimation accuracy.
	\item A two-stage coarse-to-fine dynamic removal approach for mitigating the impact of dynamic objects.
	\item Adaptive keyframe selection and downsampling strategies to ensure meshing quality and efficiency, even for sparse-channel LiDARs.
	\item Extensive evaluations in terms of localization and mapping on five public datasets for demonstrating the superiority and applicability of our method.
\end{itemize}

\section{Related Work} \label{sec2}
Many 3D LiDAR odometry and SLAM systems demonstrate excellent localization performance and can build accurate point cloud maps. LOAM \cite{loam} extracts edge and plane features, achieving pioneering results with its scan-to-scan odometry and scan-to-map mapping module. As a derivative of LOAM \cite{loam}, F-LOAM \cite{f-loam} uses a non-iterative two-stage distortion compensation method to reduce the computational cost without loss of accuracy. Using the two-stage Fast G-ICP \cite{vg-icp}, DLO \cite{dlo} directly processes the raw point cloud and delivers impressive performance. KISS-ICP \cite{kiss-icp} represents recent advancements, aiming at improving the simplicity and versatility of LiDAR odometry. Although these methods achieve reliable odometry, they only produce discrete point cloud maps, which ignore structured information and can not be directly used for navigation.

Suma++ \cite{suma++} and MARS \cite{mars} introduce a surfel-based map to represent the large-scale environments and perform real-time data association of the current scan to the surfel maps. NDT-LOAM \cite{ndt-loam} and LiTAMIN2 \cite{litamin2} use ellipsoid as basic elements, achieving efficient state estimation with corresponding map representations. VoxelMap \cite{voxelmap} proposes an effective adaptive voxel mapping method to achieve a probabilistic representation of the surrounding environment. However, these maps are relatively cluttered, with gaps between elements. Octomap \cite{octomap} and UFOMap \cite{ufomap} represent the map as occupancy grid, which can mitigate the impact of dynamic objects and be directly used for navigation, but they ignore map details. Based on OpenVDB, VDBFusion \cite{vdbfusion} offers a flexible and efficient map representation using TSDF, but its performance and accuracy depend on the chosen voxel resolution.

Some of the methods use triangular mesh for map representation and show promising performance. A widely adopted approach for constructing mesh involves fusing deep data into a TSDF map and subsequently extracting the implicit mesh on demand using Marching Cubes algorithms \cite{marching-cube}. Typical implementations of this pipeline include KinectFusion \cite{kinectfusion} and Voxblox \cite{voxblox}. Based on the planar Delaunay refinement algorithm, Li et al. \cite{sm-w-cc} produces a high quality mesh with curvature convergence. These methods, primarily designed for indoor and small-scale mesh reconstruction, have proven to be computationally-intensive. This complexity hinders their deployment in real-time applications that require processing large volumes of LiDAR scans.

Recently, the latest solutions have enabled reliable large-scale outdoor meshing. Schöps et al. \cite{surfelmeshing} fuse depth measurements in a dense surfel cloud, asynchronously triangulating smooth surface mesh. However, the mesh is not guarantee to be manifold and may also contain holes if it is not sufficiently smooth. Piazza et al. \cite{rt-cpumesh} uses the Delaunay triangulation \cite{delaunay} to reconstruct a manifold mesh in real time through single core CPU processing. Puma \cite{puma} reconstructs Poisson surfaces from a series of past scans within a sliding window and uses ray-casting for point-to-mesh registration. However, this method cannot incrementally update the global mesh map and fails to achieve real-time performance. Shine-Mapping \cite{shine-mapping} employs deep learning to achieve large-scale implicit meshing, but it requires extensive training and cannot operate online. ImMesh \cite{immesh} executes mesh partition through pull, submit, and push steps to incrementally reconstruct the triangular mesh. SLAMesh \cite{slamesh} performs GP prediction, providing robust input for meshing and enabling high-precision odometry. Mesh-LOAM \cite{mesh-loam} proposes a efficient mesh-based real-time LiDAR odometry and mapping method through implicit reconstruction and parallel spatial-hashing schemes. However, it is worth noting that these methods neglect the impact of dynamic objects on odometry and meshing. In contrast, our approach not only improves localization accuracy but also filters dynamic objects through a two-stage coarse-to-fine dynamic removal approach, achieving accurate, dense, and large-scale meshing.

\section{The Proposed Method} \label{sec3}

\subsection{System Overview} \label{sec3a}
\begin{figure*}[tbp]
	\centering
	\includegraphics[width=\textwidth]{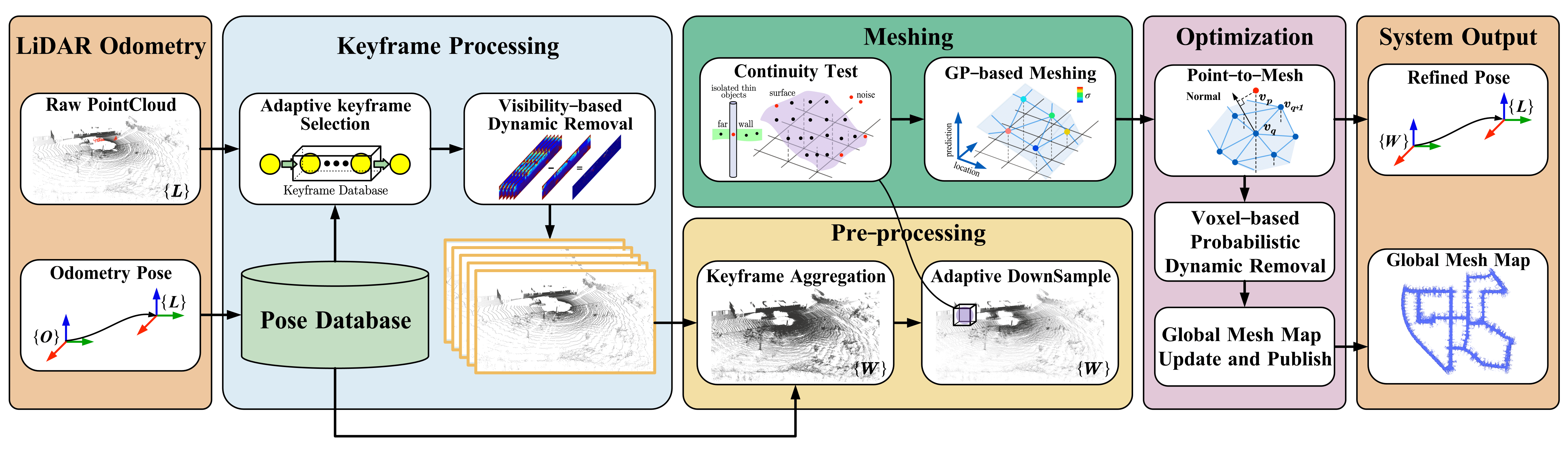}
	\caption{The system is divided into four modules. The keyframe processing module is designed for selecting keyframes and performing coarse dynamic removal to ensure registration quality. In the pre-processing module, keyframes within the sliding window are aggregated, followed by adaptive downsampling to obtain more uniform point clouds. The meshing module conducts continuity test to exclude outliers and leverage GP for mesh construction. Finally, the optimization module performs point-to-mesh registration and fine dynamic removal using voxel-based probabilistic method, obtaining the refined pose estimation and generating a continuous, accurate, and static global mesh map.}
	\label{figure2}
\end{figure*}

Fig.~\ref{figure2} illustrates the overall system architecture. As a mapping module in SLAM, the system receives the raw point $\mathcal{P}^\mathcal{L}_k$ in the LiDAR coordinate system $\mathcal{L}$ at time $k$, and the pose transformation $_\mathcal{O} \mathcal{T}^\mathcal{L}_k$ from the LiDAR coordinate system $\mathcal{L}$ to the global coordinate system $\mathcal{O}$ estimated by the odometry as the system input. The keyframe, dropped by the proposed adaptive selection mechanism, is added to the database after visibility-based coarse dynamic removal. The keyframes within the sliding window are subsequently aggregated and converted to the world coordinate system $\mathcal{W}$, then uniformly sampled by the adaptive downsampling strategy to enhance system efficiency. Continuity test is utilized to remove outliers and noise. The remaining points are divided into voxels, GP-based meshing is then conducted. In the optimization component, the pose estimated by the odometry is used as a prior for point-to-mesh registration, aligning the current scan to the global map and outputting the finer pose. Finally, after fine dynamic removal using the voxel-based probabilistic method, the current mesh is fused into the global mesh map for publication.

\subsection{Keyframe Processing and Pre-processing} \label{sec3b}
GP can effectively recover local surfaces from noisy and sparse point clouds within voxels, interpolating vertices in these local surfaces to achieve accurate mesh reconstruction \cite{slamesh}. However, GP still requires the raw point set as training data. If the raw point set is too sparse or affected by noise, the quality of the mesh reconstruction may be compromised. To address this issue, we propose the following methods.

1) Adaptive KeyFrame Selection: We use a sliding window to accumulate a certain number of keyframes as the input for meshing. Typically, a fixed threshold is used to select keyframe. However, we propose that the timing of keyframe selection may significantly impact meshing quality. Specifically, in large-scale environments, the point cloud distribution is dispersed, and points in each voxel may be insufficient or have high uncertainty, leading to lower accuracy or even failure in mesh reconstruction. Conversely, in narrow or small-scale environments, due to the denser point cloud, a larger threshold can be chosen to drop keyframe. Therefore, we scale the keyframe selection threshold based on the "spaciousness" of the instantaneous scan, defined as $m_k = \alpha m_{k-1} + \beta M_k$, where $M_k$ is the median Euclidean distance from the origin to each point, $\alpha=0.95$ and $\beta = 0.05$. We use $m_k$ as a smoothed signal to scale the keyframe translation and rotation threshold $th_k$ at time $k$, such that:

\begin{small} 
	\begin{equation} \label{eq1}
		th_k=\left\{ \begin{array}{l}
			0.0m,\ 0.0rad\ \ \ \text{if}\ m_k>20m\\
			0.3m,\ 0.1rad\ \ \ \text{if}\ m_k>10m\ \&\ m_k\le 20m\\
			0.5m,\ 0.3rad\ \ \ \text{if}\ m_k>5m\,\,\&\,\,m_k\le 10m\\
			1.0m,\ 0.5rad\ \ \ \text{if}\ m_k\le 5m\\
		\end{array} \right. 
	\end{equation}
\end{small}

According to \eqref{eq1}, keyframes are selected and added to the sliding window after coarse dynamic removal (\ref{sec3d}). The keyframes selected for the sliding window at time $k$ are defined as $\{\mathcal{F}^\mathcal{L}_{k,i}|i=1,2,...,L\}$, where $L$ is the length of the sliding window. The odometry poses corresponding to each keyframe are $\{_\mathcal{O}\mathcal{T}^\mathcal{L}_{k,i}|i=1,2,...,L\}$.

2) KeyFrame Aggregation and Adaptive Downsampling: We aggregate the keyframes within the sliding window to the current moment to serve as input, such that:
\begin{equation} \label{eq2}
	\mathcal{F}^\mathcal{L}_k = \sum_{i=1}^{l} {_\mathcal{O}\mathcal{T}^\mathcal{L}_{k,l}}^{-1} \cdot {_\mathcal{O}\mathcal{T}^\mathcal{L}_{k,i}} \cdot \mathcal{F}^\mathcal{L}_{k,i}
\end{equation}

Although the spatial distribution of points achieves greater uniformity with adaptive keyframe selection, substantial overlap remains. To address this issue, we employ an adaptive downsampling strategy for the aggregated point clouds. Following a similar rationale, if the "spaciousness" is large, a smaller downsampling size is used to preserve a higher density. Conversely, in narrow space, using a larger size can increase efficiency, such that:

\begin{small} 
	\begin{equation} \label{eq3}
		s_k=\left\{ \begin{array}{l}
			0.05m\ \ \ \text{if}\ m_k>20m\\
			0.10m\ \ \ \text{if}\ m_k>10m\ \&\ m_k\le 20m\\
			0.30m\ \ \ \text{if}\ m_k>5m\,\,\&\,\,m_k\le 10m\\
			0.50m\ \ \ \text{if}\ m_k\le 5m\\
		\end{array} \right. 
	\end{equation}
\end{small}

After processing, $\mathcal{F}^\mathcal{L}_k$ is fed into the meshing module to incrementally construct mesh maps.

\subsection{GP-based Incremental Meshing} \label{sec3c}
1) Continuity Test: Typical GP inputs all points and assumes isotropic variance, treating each point equally. However, isolated thin objects (e.g., branch, poles), noisy parts (e.g., grass, shrubs) and outliers may cause imprecise normal estimation and inconsistency between adjacent frames, affecting meshing accuracy and point-to-mesh registration quality. Therefore, before meshing, a continuity test is performed on each point obtained in \ref{sec3b} to calculate its "continuity":
\begin{equation} \label{eq4}
	\begin{split}
		c&=\omega _1\cdot c_1+w_2\cdot c_2 \\
		c_1&=\frac{1}{\left| S \right|\cdot \lVert \boldsymbol{p}_{k,i}^{\mathcal{L}} \rVert}\lVert \sum_{j\in S,j\ne i}{\left( \boldsymbol{p}_{k,i}^{\mathcal{L}}-\boldsymbol{p}_{k,\boldsymbol{j}}^{\mathcal{L}} \right)} \rVert \\
		c_2&=\frac{1}{\left| S \right|\cdot \lVert r_{k,i}^{\mathcal{L}} \rVert}\lVert \sum_{j\in S,j\ne i}{\left( r_{k,i}^{\mathcal{L}}-r_{k,\boldsymbol{j}}^{\mathcal{L}} \right)} \rVert 
	\end{split}
\end{equation}
where, $\omega_1$ and $\omega_2$ are the weights, $S$ is the set of neighbour, $\boldsymbol{p}_{k,i}^{\mathcal{L}} \in \mathbb{R}^3$ means 3D point, and $r_{k,i}$ is range of each point.

2) GP-based Meshing: Points with "continuity" less than a threshold $c_{th}$ are excluded, and the remaining points are divided into voxels for GP-based meshing. Specifically, consider a point cloud $\mathcal{F}^\mathcal{L}_{k,c}=\{p_j=(x_j,y_j,z_j,\varepsilon_j)|j=1,2.,...,n\}$ containing $n$ points $p_j$ with isotropic noise $\varepsilon_j \sim \mathcal{N}(0,\sigma_{in}^{2})$. We define that the bold lowercase letters represent vectors, and the uppercase letters represent matrices. For each 3D point, we evenly fix two of the coordinate \textit{locations} $\boldsymbol{l}=(\boldsymbol{m},\boldsymbol{n})$ to obtain a set of training points set $\boldsymbol{f}=f(\boldsymbol{l})+\boldsymbol{\varepsilon}$. The other one is called \textit{predictions} $\boldsymbol{\tilde{f}}$ with a continuous value domain. Following the Gaussian distribution, the \textit{predictions} $\boldsymbol{\tilde{f}}$ are expressed as:
\begin{equation} \label{eq5}
	\boldsymbol{\tilde{f}} = \boldsymbol{k_{mn}^{T}} (\sigma_{in}^2 \boldsymbol{I} + \boldsymbol{K_{mn}})^{-1} \boldsymbol{f}
\end{equation}
The uncertainty of \textit{predictions} is their variance:
\begin{equation} \label{eq6}
	\boldsymbol{\sigma_{n}}^2 = \boldsymbol{k_{nn}} - \boldsymbol{k_{mn}^T}(\sigma_{in}^2 \boldsymbol{I}+\boldsymbol{K_{mm}})^{-1} \boldsymbol{k_{mn}}
\end{equation}
where, $\boldsymbol{k_{nn}}$, $\boldsymbol{k_{mn}}$, and $\boldsymbol{K_{mm}}$ represent different combinations of the kernel function of \textit{locations}. We select the Gaussian kernel as the kernel function because it can represent local smooth surfaces in 2D complex manifolds.

Vertices with uncertainty $\sigma_{n}^2$ below a certain $\sigma_{match}^2$ threshold are considered accurate or valid enough to connect and build a triangle mesh, which prevents sliver mesh faces, similar to 2D Delaunay triangulation \cite{delaunay}.

In our CAD-Mesher, with fixed vertex \textit{locations}, only the 1-D \textit{predictions} require updating. When the previous $t$ data is updated below the threshold $\sigma_{update}^2$, the new \textit{predictions} can be determined using the iterative least squares method:
\begin{equation} \label{eq7}
	\tilde{f}_{t_k} = \sum_{t=1}^{t_k} (\tilde{f_t}\sigma_{t}^2) / \sum_{t=1}^{t_k} (\sigma_{t}^2),\ \text{if} \  \sigma_{t}^2 < \sigma_{update}^2
\end{equation}

We encode the index of each cell as $index = 10000000000.0 \cdot time+100000 \cdot x+1 \cdot y+0.00001 \cdot z$ and store it in a hash map. This approach ensures linear complexity for insertions, deletions, and queries. Additionally, this structure is flexible and can expand incrementally.

\begin{figure}[tbp]
	\centering
	\includegraphics[width=0.5\textwidth,height=3cm]{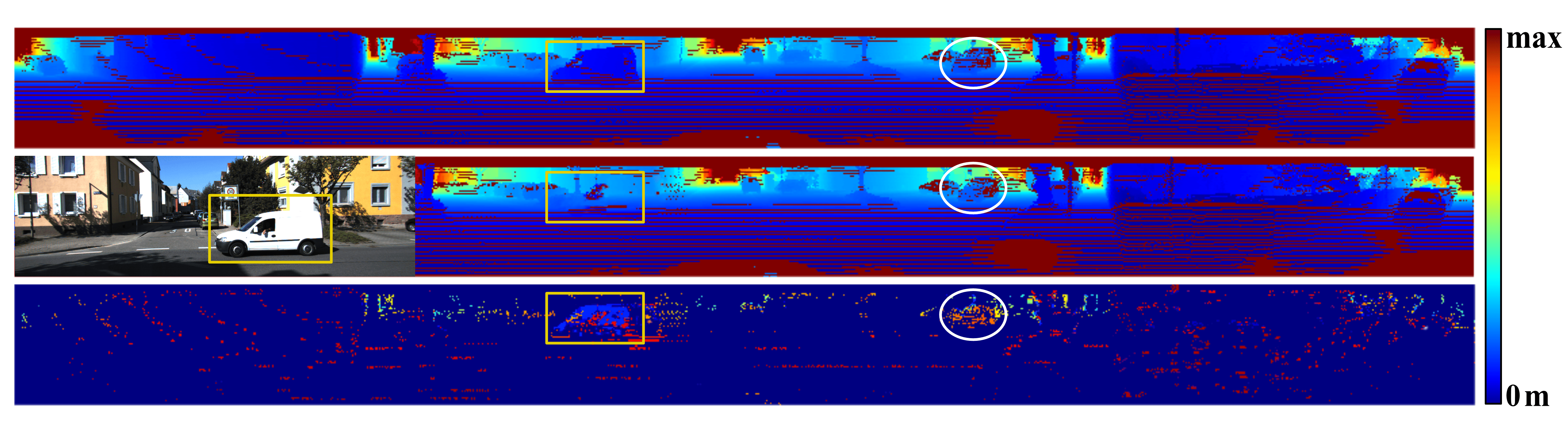}
	\caption{Coarse dynamic removal. The top row displays the range image at frame $k$. The middle row shows the range image of the aggregated point cloud at time $k-1$. The bottom row illustrates the absolute difference image between the two. The color of each pixel represents the distance value. The points in the absolute difference image is over the threshold $r_{th}$ are removed, especially for dynamic points highlighted by yellow box and white circle.}
	\label{figure3}
\end{figure}

\subsection{Two-Stage Coarse-to-Fine Dynamic Removal} \label{sec3d}
Most existing SLAM systems, whether based on point cloud or mesh, operate under the assumption of a static environment. In dynamic environments, moving objects may lead to false data associations and introduce significant ghosting into the map, which negatively impacts subsequent applications such as navigation. To address this, we propose a two-stage coarse-to-fine dynamic removal approach to effectively filter out the dynamic objects.

1) Visibility-based Coarse Dynamic Removal: Due to the keyframe accumulation, the number of points requiring processing increase. To both ensure system efficiency and mitigate the impact of dynamic objects on registration, we recommend using a visibility-based method for rapid yet coarse dynamic removal.

First, we transform the previous aggregated keyframe $\mathcal{F}^\mathcal{L}_{k-1}$ to the current time $k$. Subsequently, we perform a spherical projection on both it and the current selected keyframe $\mathcal{F}^\mathcal{L}_{k,L}$, as shown in Fig.~\ref{figure3}. Points in the absolute difference image with pixel values greater than the threshold $r_{th}$ are considered as dynamic points to be removed.

We refer to this stage as coarse dynamic removal because the performance of this method depends on the selected range image resolution. If the resolution is too large, static points will be excessively removed, reducing the quality of meshing. Conversely, too small resolution may lead to incompletely dynamic removal, affecting registration accuracy. Therefore, we choose a conservative resolution at this stage and subsequently employ a voxel-based Bayes method for the second stage to finely filter remaining dynamic points.

2) Voxel-based Probabilistic Fine Dynamic Removal: This step follows the point-to-mesh registration (\ref{sec3e}), ensuring a more accurate pose transformation. Traditional methods \cite{octomap} typically use ray-tracing to update the occupancy probability of each voxel for identifying dynamic objects. Although octree acceleration speeds up the search, this process remains time-consuming. Our approach replaces it with a method similar to the visibility check \cite{dynamicfilter}.

We first check the overlapping voxels of the current frame and the global map. The non-empty voxels are considered "occupied", whose timestamp and positions are encoded and stored as hash indexes for the voxels. Subsequently, the surrounding voxels around the robot are traversed, whose indexes are decoded to calculate the distances from them to the origin of $\mathcal{L}$. Voxels with distances less than the "occupied" voxels are considered "free". Finally, according to sensor observation $z_{1:k}$, we update the occupancy probability for each voxel $v$ using Bayesian methods:
\begin{equation} \label{eq8}
	\begin{split}
		&P\left( v|z_{1:k} \right) \\
		&=\left[ 1+\frac{P_{miss}\left( v|z_{1:k} \right)}{P_{hit}\left( v|z_{1:k} \right)}\frac{P_{miss}\left( v|z_{1:k-1} \right)}{P_{hit}\left( v|z_{1:k-1} \right)}\frac{P_0}{1-P_0} \right] ^{-1}
	\end{split}
\end{equation}
where, $P_0$ is prior probability, set to $0.5$. by using the log-odds notation, \eqref{eq8} can be rewritten as:
\begin{equation} \label{eq9}
	\begin{split}
		&Odd\left( v|z_{1:k} \right) = Odd\left( v|z_{1:k-1} \right) + Odd\left( v|z_k \right)\\
		&where,\ Odd(v) = log\left[\frac{P(v)}{1-P(v)} \right]
	\end{split}
\end{equation}

If the occupancy probability of a cell is higher than the threshold $P_{occ}$, it is considered a stable static part and will be retained in the final mesh map. Conversely, if the occupancy probability is lower than the threshold $P_{free}$, it is considered a free area, removing its content. As an alternative of ray-tracing, this method enhances computational efficiency while filtering out remaining dynamic objects. Additionally, it can be accelerated through multi-threading.

\begin{figure}[tbp]
	\centering
	\includegraphics[width=0.5\textwidth]{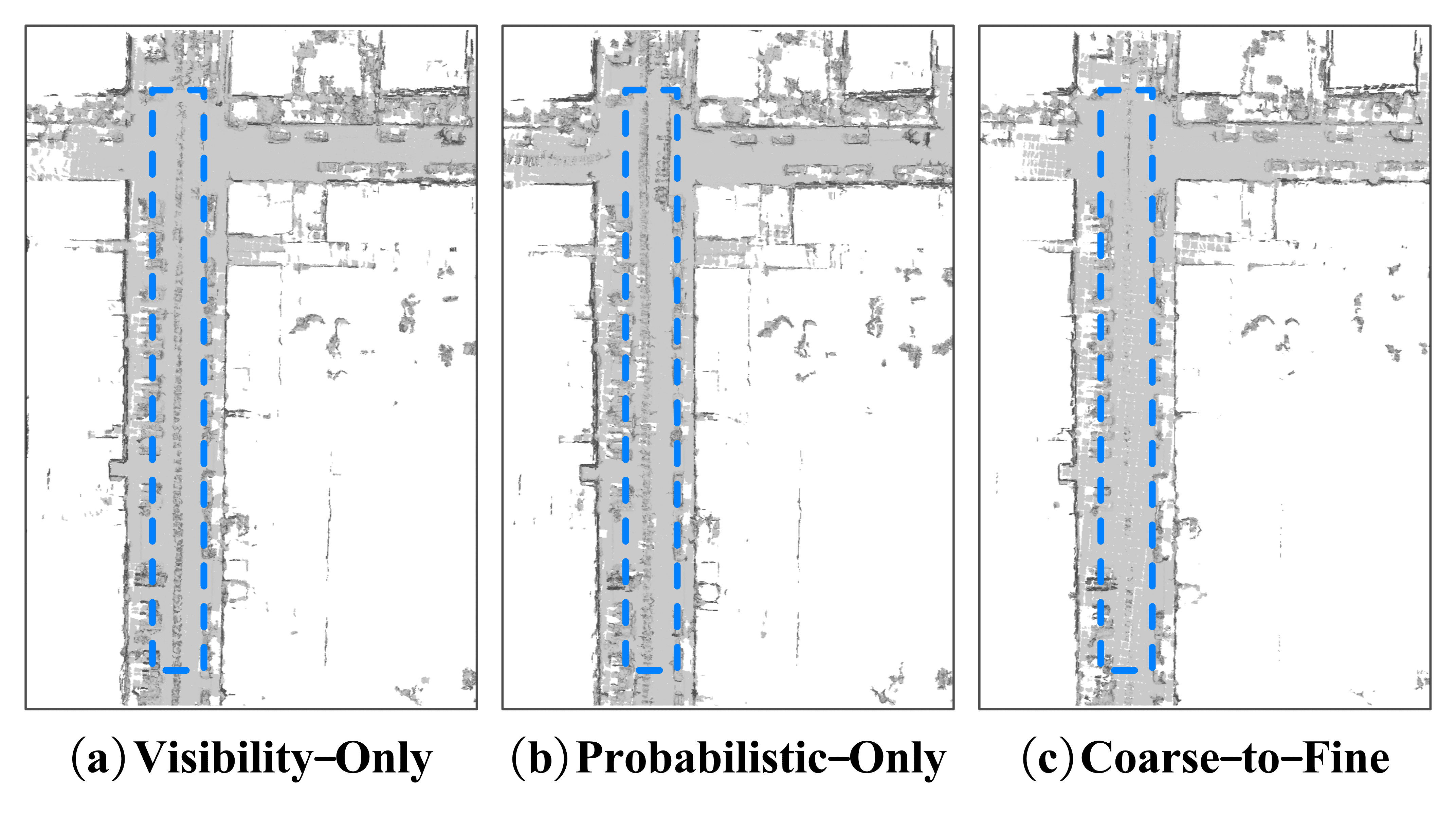}
	\caption{The effect of our dynamic removal approach. The blue dashed box indicates dynamic ghosting in the environments. (a) Using only visibility-based dynamic removal. (b) Using only voxel-based probabilistic dynamic removal. (c) our two-stage coarse-to-fine dynamic removal.}
	\label{figure4}
\end{figure}

Fig.~\ref{figure4} illustrates the effect of this step. Using a single method alone still leaves quite a few ghosting in the map. However, the proposed two-stage coarse-to-fine dynamic removal approach effectively filters out dynamic objects, resulting in a cleaner mesh map.

\subsection{Point-to-Mesh Registration  } \label{sec3e}
In contrast to the computationally-intensive ray-casting registration method \cite{puma}, we propose using nearest neighbor search for prediction vertices in the current frame for point-to-mesh registration. For each point $\boldsymbol{v}_p$ in the transformed aggregated point cloud $\mathcal{F}_{k}^{\mathcal{W}}$, we search for the nearest mesh vertex $\boldsymbol{v}_q$ in the same or in adjacent voxels. Subsequently, data association is established between $\boldsymbol{v}_p$ and the valid faces containing $\boldsymbol{v}_q$. A smooth normal $\boldsymbol{n}_q$ is computed by averaging the normals of $N_q$ adjacent vertices that are queried along the connection: 
\begin{equation} \label{eq10}
	\boldsymbol{n}_q = \frac{\sum_{q=1}^{N_q - 2}(\boldsymbol{v}_q-\boldsymbol{v}_{q-1})\times(\boldsymbol{v}_q-\boldsymbol{v}_{q+2})}{\sum_{q=1}^{N_q - 2}\lVert (\boldsymbol{v}_q-\boldsymbol{v}_{q-1})\times(\boldsymbol{v}_q-\boldsymbol{v}_{q+2}) \lVert}
\end{equation}
We construct the point-to-mesh residual using the prior $_\mathcal{O}\mathcal{T}_{k}^{\mathcal{L}}=[_\mathcal{O}\mathcal{R}_{k}^{\mathcal{L}}, _\mathcal{O}t_{k}^{\mathcal{L}}]$ estimated by LiDAR odometry:
\begin{equation} \label{eq11}
	e_p = \boldsymbol{n}_q^T \cdot (_\mathcal{O}\mathcal{R}_{k}^{\mathcal{L}} \cdot \boldsymbol{v}_p + {_\mathcal{O}t_{k}^{\mathcal{L}}} - \boldsymbol{v}_q)
\end{equation}
The optimization problem of point-to-mesh is expressed as:
\begin{equation} \label{eq12}
	\mathcal{T}=\underset{\mathcal{T}}{\text{arg}\min}\sum_{p=1}^{N_p}{e_p}
\end{equation}
We employ residual fusion and the Levenberg-Marquardt (LM) algorithm to estimate the optimal transformation. For more details, refer to SLAMesh \cite{slamesh}. Finally, we fuse the refined pose from our meshing module with the coarse pose from the odometry and publish the results at the LiDAR scanning frequency.

\section{Experiments and Results} \label{sec4}

We quantitatively evaluate meshing performance on the MaiCity simulation dataset \cite{puma} and the Newer College real-world dataset \cite{ncd} to prove the claim that our approach can build mesh maps of comparable or superior higher quality in real-time without GPU acceleration, despite of comparison with offline methods. We qualitatively meshing results on the KITTI \cite{kitti} and GroundRobot \cite{gr-loam} datasets to prove the claim that, even with sparse-channel LiDAR, our approach stably filters dynamic objects and constructs dense, consistent mesh maps. Additionally, we perform localization evaluations on the KITTI \cite{kitti} (64-beam LiDAR), UrbanLoco \cite{ul} (32-beam LiDAR), and GroundRobot \cite{gr-loam} (16-beam LiDAR) datasets to confirm the claim that our method can easily integrate with various LiDAR odometry, thereby enhancing the accuracy of pose estimation. All algorithms were executed on an Ubuntu 20.04 system with a 20-core Intel i9 2.50 GHz CPU.

\subsection{Meshing Evaluation} \label{sec4a}
We compare our method with several state-of-the-art meshing algorithms, including the TSDF-based method VDBFusion \cite{vdbfusion}, the deep learning-based offline method SHINE-Mapping \cite{shine-mapping}, and the real-time localization and meshing system SLAMesh \cite{slamesh}. For a fair comparison, we use KISS-ICP \cite{kiss-icp} to provide the input poses for VDBFusion \cite{vdbfusion} and SHINE-Mapping \cite{shine-mapping}.

1) Quantitative: We evaluate on the MaiCity dataset \cite{puma} collected by a Velodyne HDL-64E LiDAR and the Newer College dataset \cite{ncd} collected by a OS1-64 LiDAR. For the MaiCity dataset \cite{puma}, the ground truth map is a dense point cloud scanned by a high-resolution sensor. For the Newer College dataset \cite{ncd}, the ground truth is provided by officially offline-processed mesh maps. We uniformly sample the mesh maps generated by each method to form point clouds, comparing them with the ground truth, and use precision, recall, and F1-Score as evaluation metrics \cite{tat}. The evaluation results are presented in Table~\ref{table1}.

Our method achieves the highest reconstruction precision in both datasets, although its recall are not as high as that of SHINE-Mapping \cite{shine-mapping}. We attribute this discrepancy to our coarse dynamic removal method inadvertently removing some static points, and the proposed consistency test inadvertently filtering out slender poles along with noise. Nevertheless, our method achieves the highest F1-Score, demonstrating its overall effectiveness. It is also worth noting that SHINE-Mapping \cite{shine-mapping} is a deep learning-based approach for offline post-processing, which requires extensive training and cannot operate in real time. 

The signed distance error results for each method in the Newer College dataset are shown in Fig.~\ref{figure6}. It can be seen that the mesh maps constructed by our method achieves an almost error-free fit with the ground truth in the ground part. Its accuracy is comparable to the offline meshing method based on deep learning Shine-Mapping \cite{shine-mapping}, further demonstrating the effectiveness of our approach.

\begin{figure*}[tbp]
	\centering
	\includegraphics[width=\textwidth]{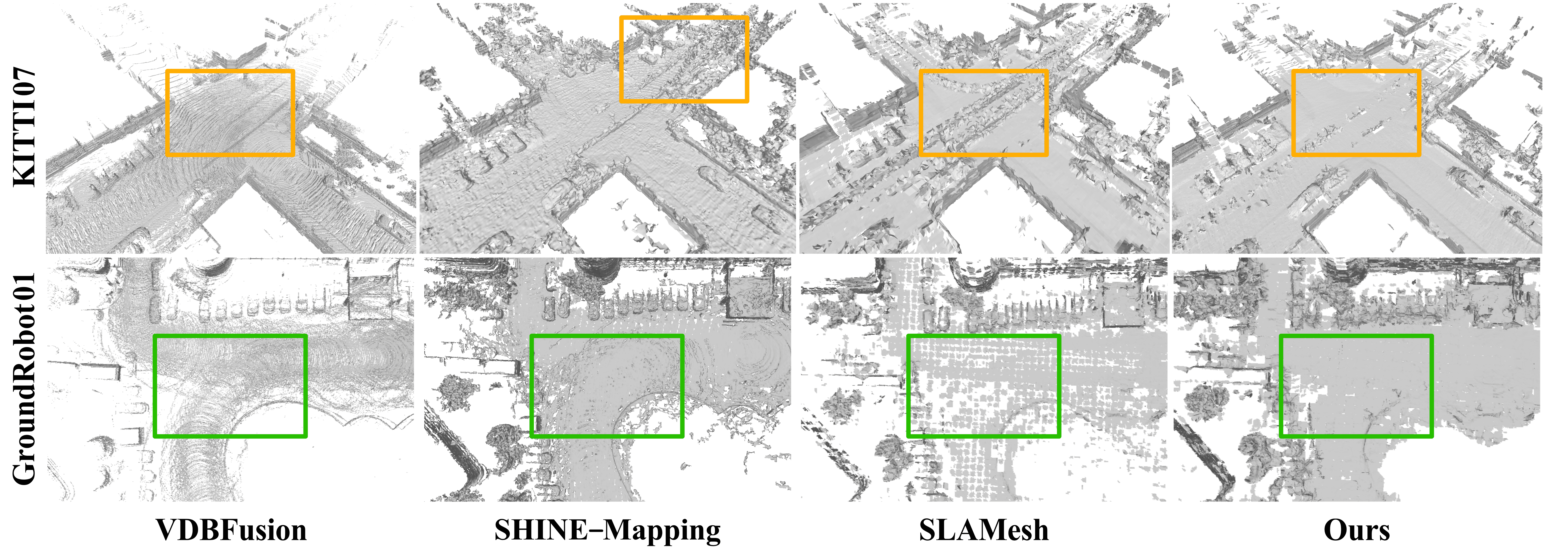}
	\caption{Qualitative results. The top row highlights the effect of our dynamic removal, with the yellow box indicating dynamic ghosting. The bottom row shows the robust performance of our method with sparse-channel LiDAR, where green boxes indicate the density or inconsistencies of meshing.}
	\label{figure5}
\end{figure*}
\begin{table}[t]
	\caption{Quantitative Comparative Results (\%) of Meshing.}
	\centering
	\scalebox{1.05}{
		\begin{tabular}{ccccc}
			\toprule
			\rule{-1pt}{8pt}
			Dataset & Method & Precision & Recall & F1-Score \\
			
			\midrule
			\rule{-1pt}{8pt}
			\multirow{4}{*}{\makecell{MaiCity \\ ($\delta = 0.1m$)}}
			& VDBFusion & 90.07 & 91.19 & 90.63 \\
			
			& SHINE-Mapping & 92.12 & \textbf{95.05} & 93.56 \\
			
			& SLAMesh & 87.69 & 82.73 & 85.86 \\
			
			& Ours & \textbf{95.50} & 94.18 & \textbf{94.84} \\
			
			\midrule
			\rule{-1pt}{8pt}
			\multirow{4}{*}{\makecell[c]{Newer \\ College \\ ($\delta = 0.2m$)}}
			& VDBFusion & 94.38 & 86.84 & 90.45 \\
			
			& SHINE-Mapping & 90.74 & \textbf{91.96} & 91.34 \\
			
			& SLAMesh & 87.13 & 87.63 & 87.38 \\
			
			& Ours & \textbf{95.52} & 89.76 & \textbf{92.55} \\
			\bottomrule
			\multicolumn{4}{l}{\footnotesize The \textbf{bold} font denotes the best results.}
		\end{tabular}
	}
	\label{table1}
\end{table}

2) Qualitative: Fig.~\ref{figure5} illustrates the performance of each method on the KITTI07 and GroundRobot01 sequences. In KITTI07, moving vehicles at intersections leave obvious ghosting in the maps for the other mesh baselines, which impedes following navigation applications. Although VDBFusion \cite{vdbfusion} migrates dynamic impact using space caving technique, rough ground and residual ghosting still exist. In contrast, our CAD-Mesher effectively filters dynamic objects and ensures map consistency through the proposed two-stage coarse-to-fine dynamic removal approach. However, few dynamic remnants are still observed in the map, likely related to the chosen resolution and voxel size.

\begin{figure}[tbp]
	\centering
	\includegraphics[width=0.5\textwidth]{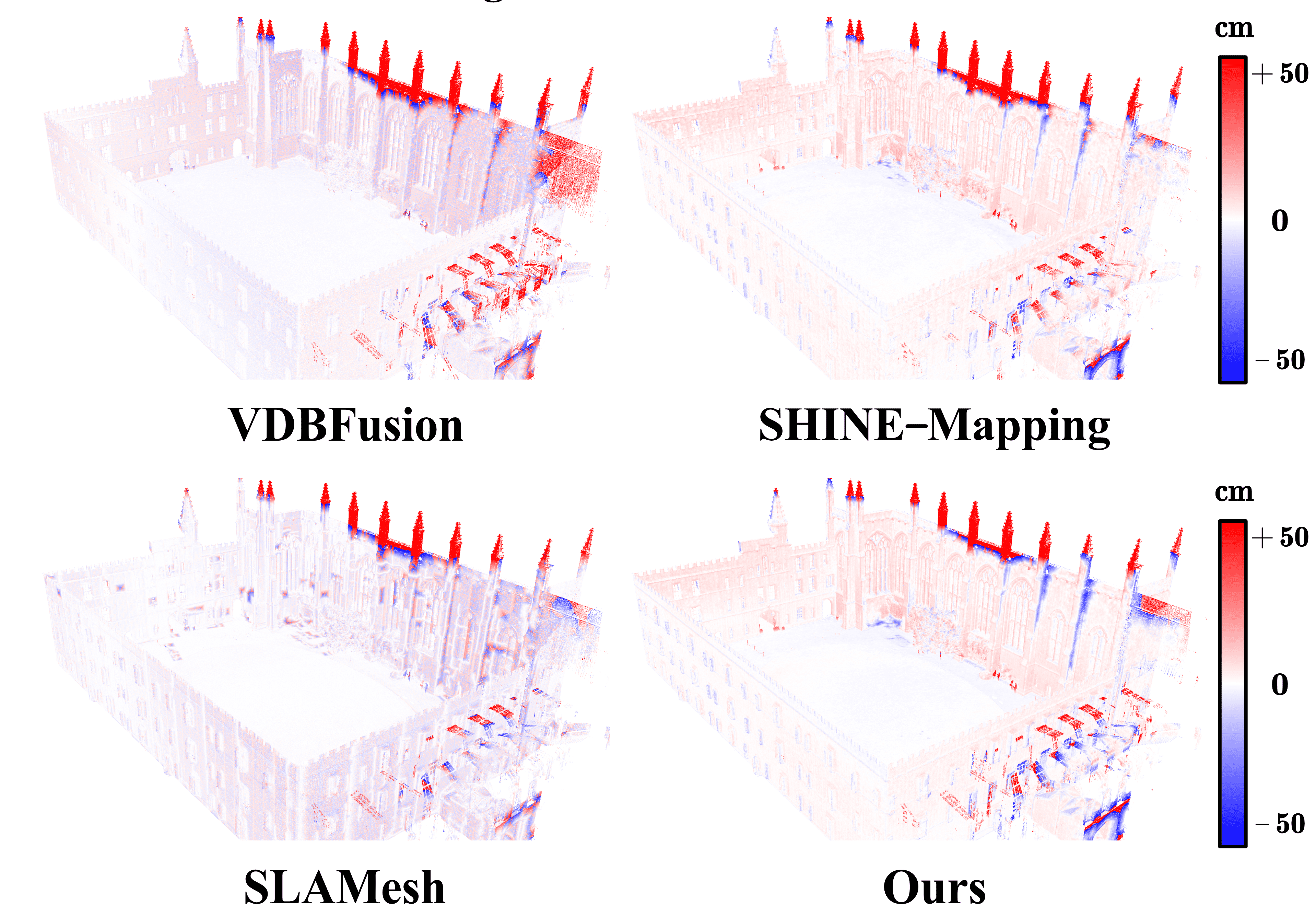}
	\caption{Signed distance error visualization of all comparison methods on the Newer College dataset \cite{ncd}. The red color represents the positive distance between the resulting mesh and the ground truth, while the blue color indicates the negative distance. The brighter the color, the greater the error is.}
	\label{figure6}
\end{figure}

In the GroundRobot01 sequence, the sparse-channel LiDAR presents a challenge to registration accuracy and meshing quality. Due to the sparsity of the point cloud, both VDBFusion \cite{vdbfusion} and SLAMesh \cite{slamesh} produce many holes in the ground, compromising the continuity of mesh map. Although SHINE-Mapping \cite{shine-mapping} mitigates the impact of sparsity and generates a dense map, it exhibits stratification in the green box due to odometry drift. However, our method provides more accurate poses for meshing by refining pose estimation of odometry, thereby ensuring consistency of meshing.

\begin{table*}[ht]
	\caption{Localization Accuracy Evaluation Using APE ($m$) on KITTI, UrbanLoco, and GroundRobot datasets}
	\centering
	\scalebox{0.9}{
		\begin{tabular}{ccccccccc}
			\toprule
			\rule{-1pt}{8pt}
			\multirow{2}{*}{\textbf{\normalsize{Dataset}}} & \multicolumn{3}{c}{\textbf{KITTI}} & \multicolumn{3}{c}{\textbf{UrbanLoco}} & \multicolumn{2}{c}{\textbf{GroundRobot}} \\
			
			\cmidrule(r){2-4} \cmidrule(r){5-7} \cmidrule(r){8-9}
			\rule{-1pt}{6pt}
			
			& KITTI00 & KITTI05 & KITTI07 & UrbanLoco01 & UrbanLoco03 & UrbanLoco05 & GroundRobot01 & GroundRobot02 \\
			\midrule
			
			LiDAR channel & 64 & 64 & 64 & 32 & 32 & 32 & 16 & 16 \\
			\midrule
			
			Characteristics & large-scale & low dynamic & medium dynamic & high dynamic & medium dynamic & high dynamic & sparse-channel & sparse-channel \\
			\midrule
			
			A-LOAM & 6.76/8.07 & 3.64/4.19 & 0.67/0.70 & 1.50/1.83 & 1.84/2.15 & 10.17/15.19 & \textbf{1.06}/\textbf{1.34} & 4.64/4.96 \\
			\midrule
			
			DLO & 5.52/6.67 & 3.00/3.34 & 1.46/1.54 & 1.70/2.03 & \textbf{1.61}/\textbf{1.83} & 1.58/1.72 & 1.31/1.37 & 2.47/2.63 \\
			\midrule
			
			KISS-ICP & 7.43/8.55 & 2.83/3.32 & 0.64/0.67 & 2.94/4.16 & 2.70/3.09 & 1.68/1.96 & 1.76/2.00 & 3.94/4.27 \\
			\midrule
			
			SLAMesh & 4.46/5.89 & 2.03/2.42 & 0.96/1.04 & 1.18/1.37 & 1.89/2.23 & 1.62/1.78 & 1.73/2.08 & 3.03/3.32 \\
			\midrule
			
			RF-A-LOAM & 6.81/8.05 & 3.12/3.55 & 0.69/0.71 & 1.44/1.74 & 1.81/2.11 & 1.45/1.62 & 1.12/1.40 & 4.77/5.13 \\
			\midrule
			\midrule
			
			A-LOAM+Ours & 5.94/6.77 & 3.51/4.31 & \textbf{\textcolor{blue}{0.59}}/\textbf{\textcolor{blue}{0.61}} & \textbf{\textcolor{blue}{1.02}}/\textbf{\textcolor{red}{1.14}} & \textbf{\textcolor{blue}{1.57}}/\textbf{\textcolor{blue}{1.80}} & \textbf{1.30}/1.47 & \textbf{\textcolor{red}{0.78}}/\textbf{\textcolor{red}{0.83}} & \textbf{2.20}/\textbf{2.30} \\
			\midrule
			
			DLO+Ours & \textbf{4.43}/\textbf{5.06} & 2.58/2.85 & 0.95/1.02 & \textbf{\textcolor{red}{1.01}}/\textbf{\textcolor{blue}{1.16}} & \textbf{\textcolor{red}{1.32}}/\textbf{\textcolor{red}{1.56}} & \textbf{\textcolor{red}{1.22}}/\textbf{\textcolor{red}{1.31}} & \textbf{\textcolor{blue}{0.82}}/\textbf{\textcolor{blue}{1.02}} & \textbf{\textcolor{blue}{1.90}}/\textbf{\textcolor{red}{1.98}} \\
			\midrule
			
			KISS-ICP+Ours & 5.25/6.50 & \textbf{2.01}/\textbf{2.37} & \textbf{\textcolor{red}{0.52}}/\textbf{\textcolor{red}{0.60}} & 1.21/1.45 & 1.69/1.89 & \textbf{\textcolor{blue}{1.23}}/\textbf{\textcolor{blue}{1.36}} & 1.43/1.61 & 2.31/2.38 \\
			\midrule
			
			SLAMesh+Ours & \textbf{\textcolor{red}{2.33}}/\textbf{\textcolor{red}{2.59}} & \textbf{\textcolor{red}{1.22}}/\textbf{\textcolor{red}{1.41}} & 0.79/0.82 & \textbf{1.15}/\textbf{1.35} & 1.71/1.92 & 1.34/\textbf{1.46} & 1.48/1.60 & 2.23/2.32 \\
			\midrule
			
			Ours & \textbf{\textcolor{blue}{2.95}}/\textbf{\textcolor{blue}{/3.53}} & \textbf{\textcolor{blue}{1.38}}/\textbf{\textcolor{blue}{1.61}} & \textbf{0.62}/\textbf{0.66} & 1.26/1.48 & 1.72/1.92 & 1.36/1.49 & 1.20/1.80 & \textbf{\textcolor{red}{1.89}}/\textbf{\textcolor{blue}{1.99}}  \\
			\bottomrule
			\multicolumn{9}{l}{\footnotesize All errors are represented as mean (m) / rmse (m) (the smaller the better). \textbf{\textcolor{red}{Red}}, \textbf{\textcolor{blue}{blue}} and \textbf{bold} fonts denote the first, second and third place, respectively.}
		\end{tabular}
	}
	\label{table2}
\end{table*}

\subsection{Localization Evaluation} \label{sec4b}
\textbf{Experiment Setup}: We select several typical sequences from the KITTI \cite{kitti}, UrbanLoco \cite{ul}, and GroundRobot \cite{gr-loam} datasets for localization evaluation. The LiDAR channels used in these sequences and their characteristics are summarized in Table.~\ref{table2}. We evaluate the following LiDAR odometry systems using the mean and RMSE of Absolute Pose Error (APE) as metrics: (i) Feature-based two-stage LO: A-LOAM \cite{loam}; (ii) State-of-the-art direct LO: DLO \cite{dlo}, KISS-ICP \cite{kiss-icp}; (iii) State-of-the-art mesh-based LO: SLAMesh \cite{slamesh}; (iv) According to RF-LIO \cite{rf-lio}, our own-reimplemented dynamic LO solution: RF-A-LOAM\footnotemark{}. (v) For comparison, our basic system (Ours: the bottom line of Table.~\ref{table2}) is CAD-Mesher considering the constant velocity motion model as pose prior.

\textbf{Evaluation Setup}: (i) We integrate our module with A-LOAM \cite{loam} (replace its mapping module), DLO \cite{dlo}, KISS-ICP \cite{kiss-icp}, and SLAMesh \cite{slamesh}, generating their variants: A-LOAM+Ours, DLO+Ours, KISS-ICP+Ours, SLAMesh+Ours. We then compare these baselines with their variants to prove the claim that our plug-and-play CAD-Mesher module can effectively improve the accuracy of LiDAR odometry; (ii) To prove that the superiority of our method in terms of localization, we conduct comparison experiments between A-LOAM \cite{loam}, DLO \cite{dlo}, KISS-ICP \cite{kiss-icp}, SLAMesh \cite{slamesh}, RF-A-LOAM and Ours.

\textbf{Experiment Results}: The results are shown in Table.~\ref{table2}, and we can draw the following conclusions:

1) After integrating with our CAD-Mesher module, the localization accuracy of various LiDAR odometry systems is further improved, generally outperforming the origin systems. Interestingly, while KISS-ICP \cite{kiss-icp} is inferior to DLO \cite{dlo} in the KITTI00 sequence, KISS-ICP+Ours outperforms DLO \cite{dlo}, which is a common phenomenon for other baselines.

2) In highly dynamic scenarios, LiDAR odometry integrating with our meshing module can form a more robust system and achieve better localization accuracy. Especially, A-LOAM \cite{loam} originally fails in the UrbanLoco05 sequence, but A-LOAM+Ours gains significant performance improvement, surpassing the most of methods. We attribute this improvement to our two-stage coarse-to-fine dynamic removal approach. Additionally, our method handles dynamic situations more effectively compared with dynamic LO solution RF-A-LOAM.

3) Even without prior pose estimated by LiDAR odometry, our method using the constant velocity motion model achieves competitive performance, outperforming against other LiDAR odometry, which demonstrates the effect of the proposed point-to-mesh registration.

4) It is also interesting to note that point cloud registration-based odometry integrating with our meshing module generally results in greater improvements than mesh-based odometry integrating with our CAD-Mesher. We speculate that point cloud-based registration is decoupled from our point-to-mesh registration and presents a complementary effect.

In summary, our CAD-Mesher mapping module can seamlessly integrate with various LiDAR odometry systems to further improve localization accuracy. Additionally, the integrated system can effectively cope with highly dynamic scenes and sparse-channel LiDAR data.

\begin{table}[t]
	\caption{Ablation Results. Meshing evaluation (\%) on the MaiCity dataset. Localization evaluation using RMSE (m) of APE on the UrbanLoco05 sequence.}
	\centering
	\scalebox{0.85}{
		\begin{tabular}{c|cccc|ccc|c}
			\toprule
			\rule{-1pt}{8pt}
			
			Variant & D.R. & C.T. & A.D. & S.W. & Precision & Recall & F1-Score & APE\\
			\midrule
			\rule{-1pt}{8pt}
			
			(a) & \usym{2714} & \usym{2714} & \usym{2714} & 1 & 93.60 & 83.00 & 87.98 & 1.71\\
			
			(b) & \usym{2714} & \usym{2714} & \usym{2714} & 3 & 95.27 & 89.77 & 92.44 & 1.55\\
			
			(c) & \usym{2714} & \usym{2714} & \usym{2714} & 5 & \underline{\textbf{96.32}} & \underline{94.18} & \underline{94.84} & \underline{\textbf{1.36}} \\
			
			(d) & \usym{2714} & \usym{2714} & \usym{2714} & 8 & 96.21 & \textbf{94.36} & \textbf{95.27} & 1.42 \\
			
			(e) & \usym{2714} & \usym{2717} & \usym{2714} & 5 & 95.21 & 94.29 & 94.75 & 1.62 \\
			
			(f) & \usym{2714} & \usym{2714} & \usym{2717} & 5 & 95.16 & 94.78 & 94.97 & 1.41 \\
			
			(g) & \usym{2717} & \usym{2714} & \usym{2714} & 5 & 95.72 & 94.67 & 95.19 & 1.66 \\
			\bottomrule
			\multicolumn{9}{l}{\footnotesize The \textbf{bold} font denotes the best results. The \underline{underscore} denotes the default setting.}
		\end{tabular}
	}
	\label{table3}
\end{table}

\subsection{Ablation Study} \label{sec4c}
\begin{figure}[tbp]
	\centering
	\includegraphics[width=0.5\textwidth]{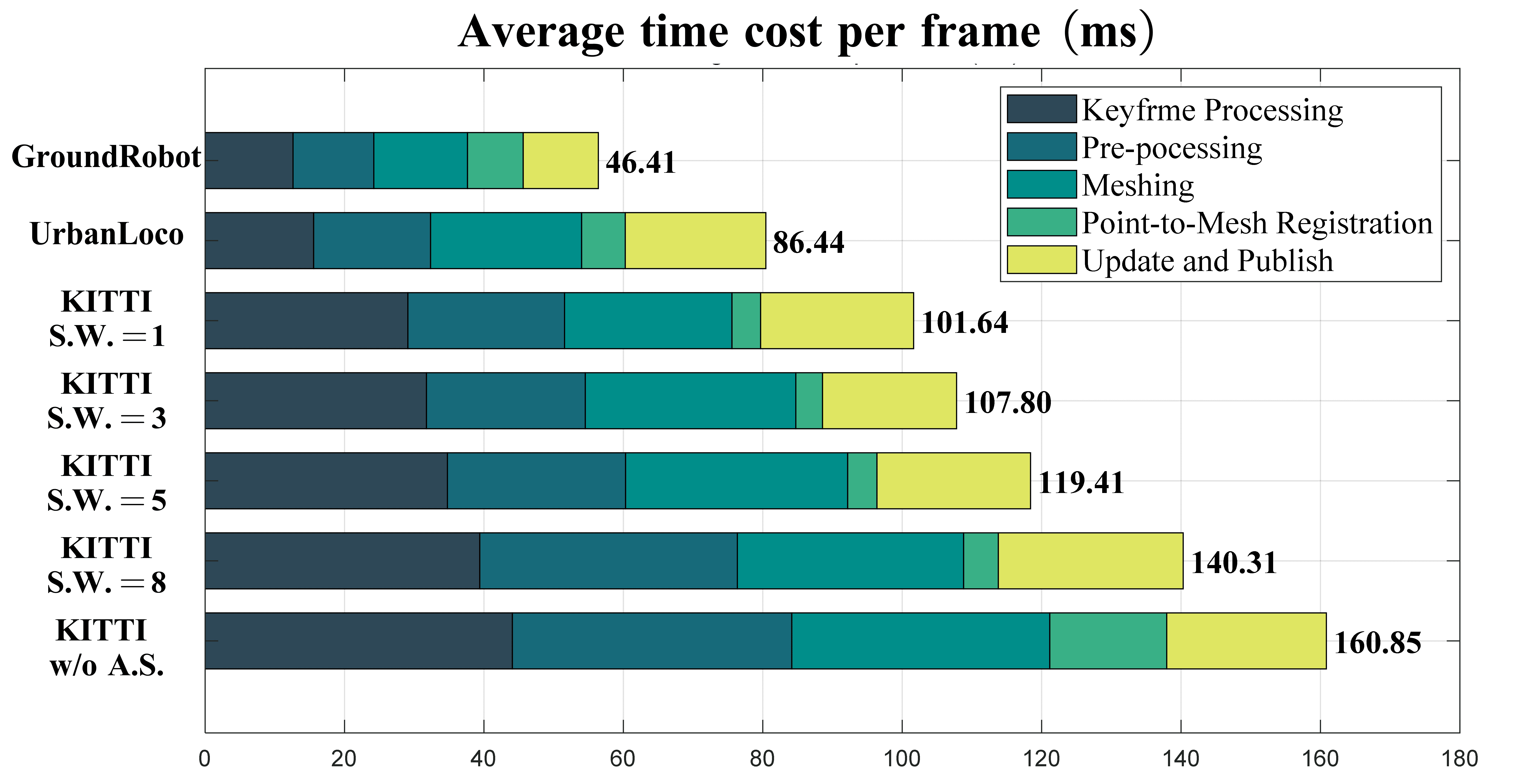}
	\caption{The time cost for each component in different datasets and variants.}
	\label{figure7}
\end{figure}

We conduct ablation studies to investigate the contribution of each component to our system, including Dynamic Removal (D.R.), Continuity Test (C.T.), Adaptive Downsampling (A.D.), and Sliding Window size(S.W.). In the MaiCity dataset \cite{puma}, we assess map quality using precision, recall, and F1-Score as evaluation metrics. Notably, since the MaiCity dataset \cite{puma} does not contain dynamic objects, we perform an additional localization evaluation using the RMSE of APE on the UrbanLoco05 sequence to demonstrate the performance of our dynamic removal approach. For each variant, we use KISS-ICP \cite{kiss-icp} as the odometry input. The results are presented in Table.~\ref{table3} and Fig.~\ref{figure7}.

According to (a)-(d), as the size of the sliding window increases, both the completion and F1-Score of meshing improve, which demonstrates the superiority of our keyframe and sliding window method. However, if the sliding window size is excessively increased, although meshing quality slightly improves, the computational time rises significantly. We consider the improvement is marginal compared to the significant increase in runtime. Additionally, localization accuracy tends to decrease due to potential interference from the increased noise.

Comparing (c), (e), and (g), the employments of continuity test and dynamic removal modules enhance the accuracy of localization. Although this may lead to a slight decrease in meshing quality due to occasionally inadvertent deletions of static points, we devote to implement a convenient, accurate, and dense mesh-based mapping module in SLAM, which needs to balance localization accuracy and meshing quality, different from an offline mesh generator.

Finally, according to (c), (f), and Fig.~\ref{figure7}, employing adaptive downsampling strategy reduces the system time cost by nearly half while maintaining the meshing quality. This highlights the advantage of our adaptive downsampling mechanism in terms of time efficiency.

\subsection{Time Analysis} \label{sec4d}
Figure~\ref{figure7} summarizes the time consumption of our approach across different datasets. While typical mapping modules usually require an operating frequency above 1 Hz \cite{loam}, our system achieves an operating efficiency of 8 Hz with a 64-beam LiDAR, meeting real-time requirements of LiDAR odometry system. Additionally, our approach can operate at frequencies over 10 Hz with 32- or 16-beam LiDAR, demonstrating its high efficiency.

\section{Conclusion}
In this paper, we propose a plug-and-play meshing module that can easily integrate with various LiDAR odometry systems to achieve more accurate pose estimation and construct consistent, dense, and clean maps. The core components of our module include the sliding window-based keyframe aggregation and the two-stage coarse-to-fine dynamic removal approach, which benefits for high-quality mesh reconstruction. Experiments on multiple publicly available datasets validate the effectiveness of our approach. 

\bibliographystyle{IEEEtran}
\bibliography{myRef.bib}

\begin{IEEEbiography}[{\includegraphics[width=1in,height=1.25in,clip,keepaspectratio]{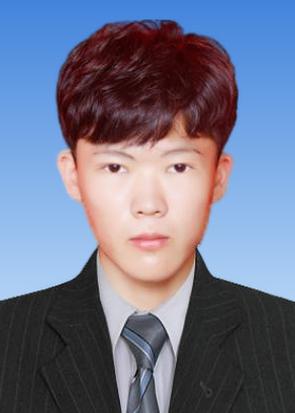}}]{Yanpeng Jia}
	received the B.S. degree in Department of Mechanical Engineering from Harbin Institute of Technology, Weihai, China in 2023. He is currently pursuing the M.S degree with the Shenyang Institute of Automation, Chinese Academy of Sciences. His current research interests include robot control, LiDAR Odometry and dynamic SLAM.
\end{IEEEbiography}
\vspace{-7mm}
\begin{IEEEbiography}[{\includegraphics[width=1in,height=1.25in,clip,keepaspectratio]{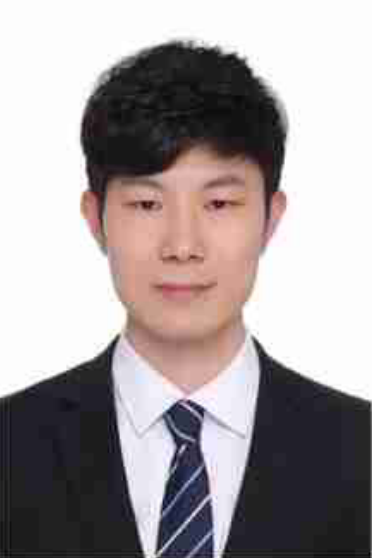}}]{Fengkui Cao}
	received the bachelor’s degree in vehicle engineering from Shandong Agricultural University, Shandong, China, in 2013, and the Ph.D. degree in control science and engineering from the Dalian University of Technology, Dalian, China, in 2020. He is currently working in the State Key Laboratory of Robotics at Shenyang Institute of Automation, Shenyang, China. His research interests include unmanned ground vehicles’autonomous navigation and long-term localization, SLAM, and robotic vision.
\end{IEEEbiography}
\vspace{-7mm}
\begin{IEEEbiography}[{\includegraphics[width=1in,height=1.25in,clip,keepaspectratio]{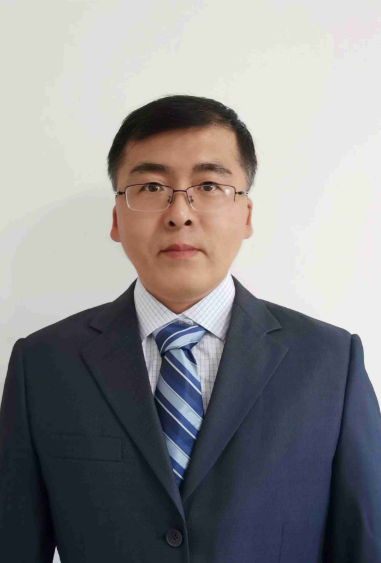}}]{Ting Wang}
	received the B.S degree in Department of Automation from Dalian University of Technology in 2001. In 2007, he graduated from Shenyang Institute of Automation, majoring in pattern recognition and intelligent systems, and obtained the Ph.D. degree.Since then he has been working in the State Key Laboratory of Robotics at Shenyang Institute of Automation. His research interests include robot control,special robot technology and pattern recognition and intelligent system.
\end{IEEEbiography}
\vspace{-7mm}
\begin{IEEEbiography}[{\includegraphics[width=1in,height=1.25in,clip,keepaspectratio]{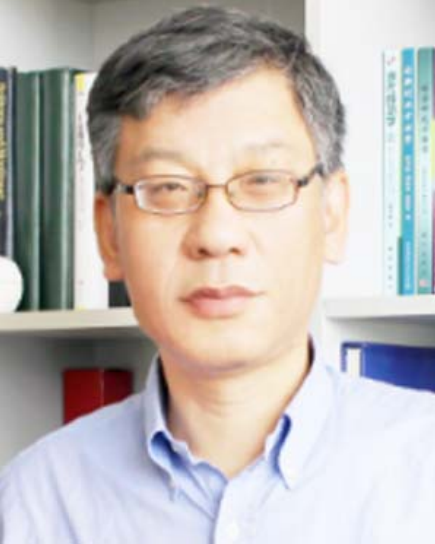}}]{Yandong Tang}
	received B.S. and M.S. degrees from the Department of Mathematics, Shandong University, Jinan, China, in 1984 and 1987, and the Ph.D. degree in applied mathematics from the University of Bremen, Bremen, Germany, in 2002. He is currently a Professor with the Shenyang Institute of Automation, Chinese Academy of Sciences, Beijing, China. His current research interests include robot vision, pattern recognition, and numerical computation.
\end{IEEEbiography}
\vspace{-7mm}
\begin{IEEEbiography}[{\includegraphics[width=1in,height=1.25in,clip,keepaspectratio]{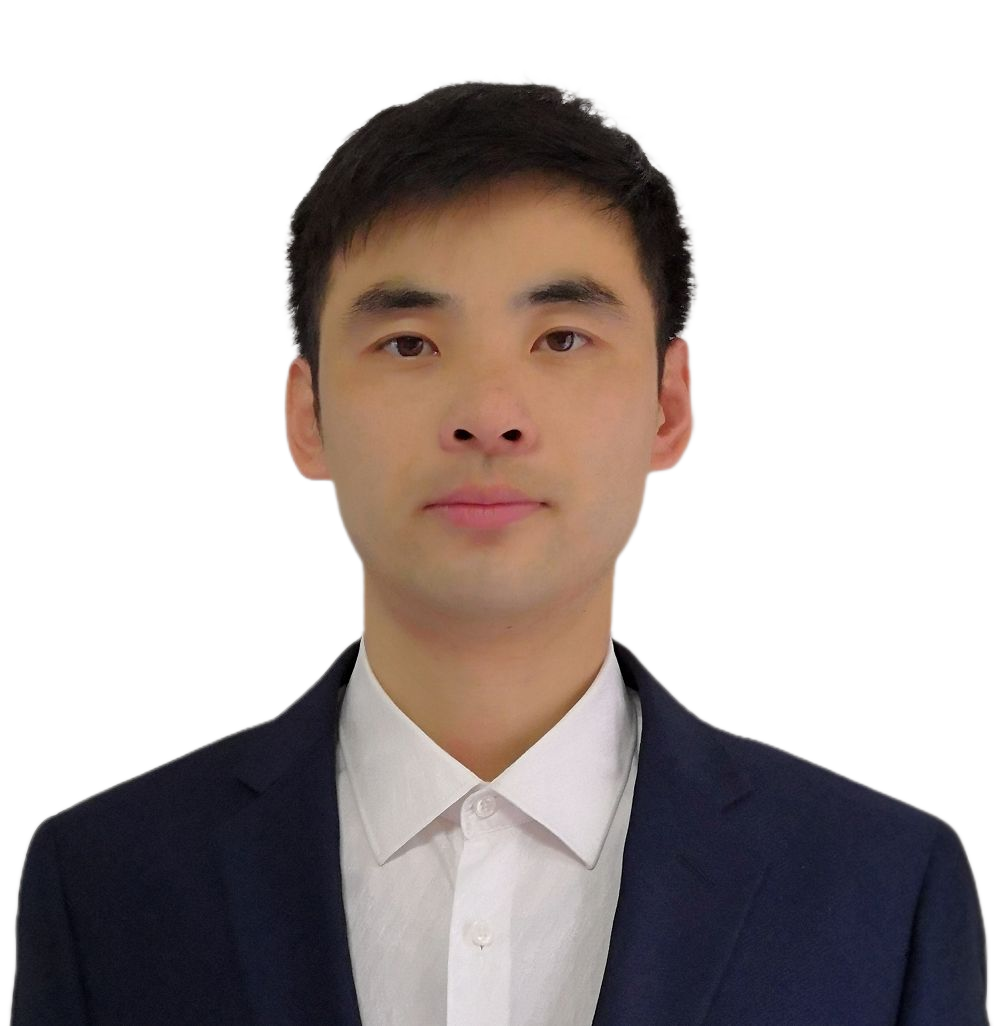}}]{Shiliang Shao}
	received the B.S degree in Department of Electronic Information Engineering from Southwest University in 2011, and received the M.S degree in Department of Information Science and Engineering from Northeast University in 2013. Since then, he has been working in State Key Laboratory of Robotics at Shenyang Institute of Automation. From 2016 to 2020, he is studying for a Ph.D. at Northeastern University. He received the Ph.D. degree in 2020. His research interests include special robot technology, physiological signal analysis, and intelligent systems.
\end{IEEEbiography}
\vspace{-7mm}
\begin{IEEEbiography}[{\includegraphics[width=1in,height=1.25in,clip,keepaspectratio]{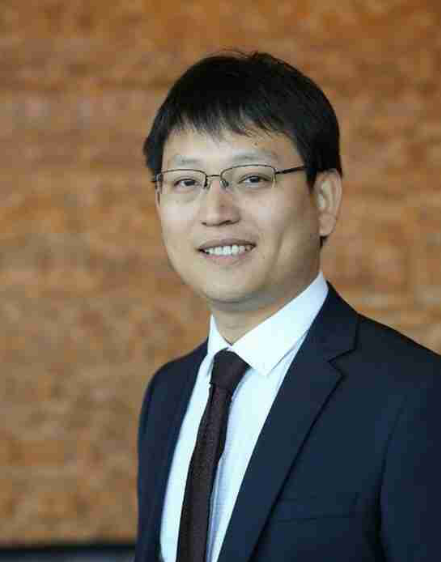}}]{Lianqing Liu}
	(Senior Member, IEEE) received the B.S. degree in industry automation from Zhengzhou University, Zhengzhou, China, in 2002, and the Ph.D. degree in pattern recognition and intelligent systems from the Shenyang Institute of Automation, Chinese Academy of Sciences, Shenyang, China, in 2009. He is currently a Professor with the Shenyang Institute of Automation, Chinese Academy of Sciences. His current research interests include micro/nanorobotics, biosyncretic robotics, and intelligent control.
\end{IEEEbiography}

\vfill

\end{document}